\documentclass{article}
\usepackage{spconf,amsmath,graphicx}
\usepackage{amssymb}
\usepackage{times}
\usepackage{float}
\usepackage{epsfig}
\usepackage{graphicx}
\usepackage{booktabs}
\usepackage{color,colortbl}
\definecolor{mygray}{gray}{.9}
\usepackage{bbm}
\usepackage{algorithm,algorithmic}
\usepackage{enumitem}
\usepackage{flushend}
\usepackage[misc]{ifsym}
\usepackage{enumitem}

\title{Transductive CLIP with Class-Conditional Contrastive Learning}

\name{Junchu Huang$^{1,*}$, Weijie Chen$^{3,2}$, Shicai Yang$^{2}$, Di Xie$^{2}$, Shiliang Pu$^{2,}$\textsuperscript{\Letter}, Yueting Zhuang$^{3,}$\textsuperscript{\Letter}\thanks{* \, This work is done when Junchu Huang was an intern in Hikvision Research Institute. \Letter \, Corresponding Authors}}
\address{$^{1}$South China University of Technology, Guangzhou, China\\
$^{2}$ Hikvision Research Institute, Hangzhou, China\\
$^{3}$ Zhejiang University, Hangzhou, China}

\begin{document}
\maketitle
\begin{abstract}
Inspired by the remarkable zero-shot generalization capacity of vision-language pre-trained model, we seek to leverage the supervision from CLIP model to alleviate the burden of data labeling. However, such supervision inevitably contains the label noise, which significantly degrades the discriminative power of the classification model. In this work, we propose Transductive CLIP, a novel framework for learning a classification network with noisy labels from scratch. Firstly, a \emph{class-conditional contrastive learning} mechanism is proposed to mitigate the reliance on pseudo labels and boost the tolerance to noisy labels. Secondly, \emph{ensemble labels} is adopted as a pseudo label updating strategy to stabilize the training of deep neural networks with noisy labels. This framework can reduce the impact of noisy labels from CLIP model effectively by combining both techniques. Experiments on multiple benchmark datasets demonstrate the substantial improvements over other state-of-the-art methods.
\end{abstract}

\section{Introduction}
\label{sec:intro}
The revolutionized successes of deep neural networks in a variety of computer vision applications are conferred by large databases with accurate annotation \cite{deng2009imagenet,chen2019all,lin2019attribute}. In many real-world scenarios, data labeling is very costly in terms of resource and time consumption. Several efforts had been made for unsupervised model optimization in the past \cite{chen2020simple,UIC,he2020momentum,SFOD,SSNLL}. Recently, Contrastive Language-Image Pretraining (CLIP) \cite{radford2021learning} has emerged as a promising alternative for generalizing vision tasks. To alleviate the burden of data labeling, there is a strong motivation to leverage the unlabeled data supervised from the CLIP model in a transductive learning manner. However, it cannot achieve satisfactory performance by directly learning from the pseudo labels predicted by CLIP model since the classification model is prone to fit and memorize the label noise \cite{shu2019meta}, leading to the performance degeneration.
\begin{figure}[H]
  \centering
  \centerline{\includegraphics[width=8.5cm]{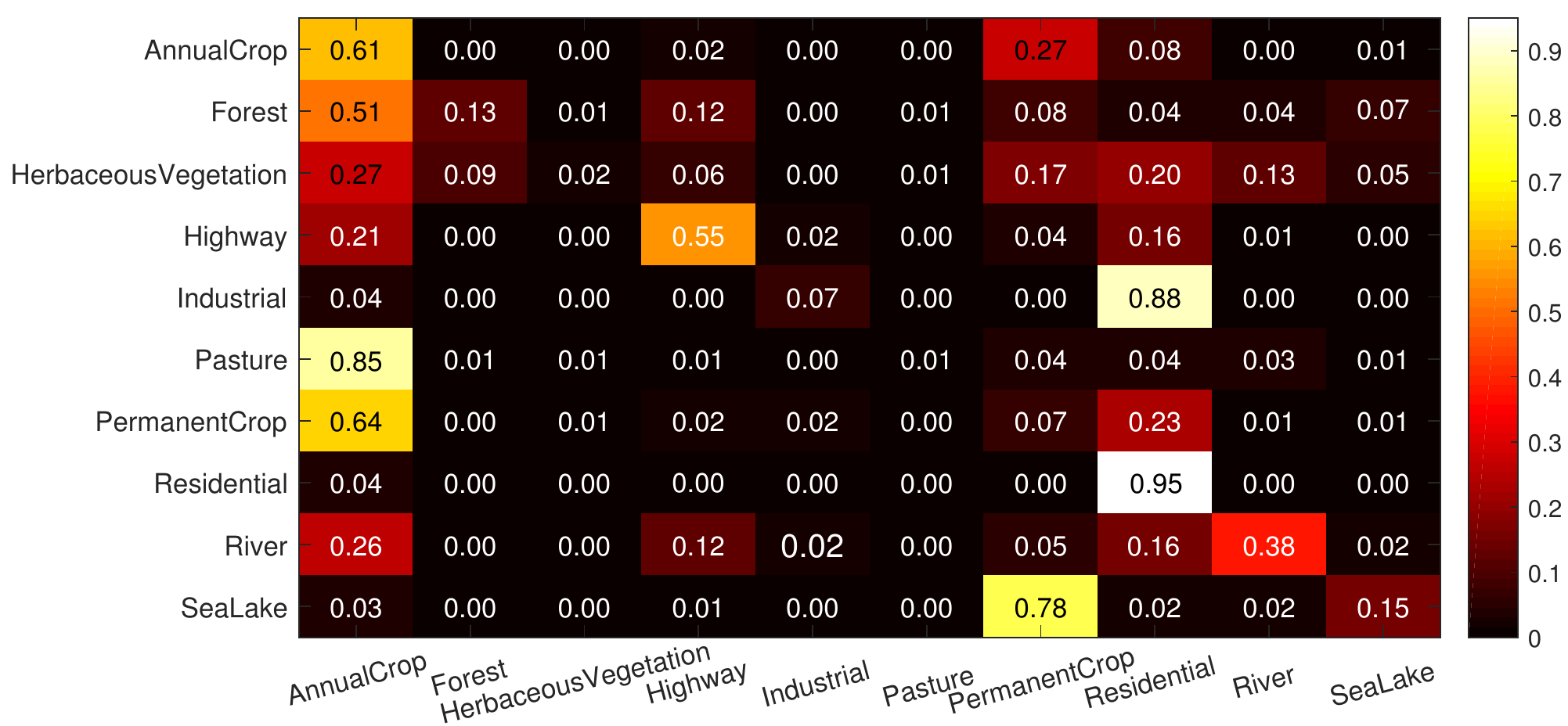}}
\vskip -0.2in
\caption{The realistic noise matrix in the EuroSAT dataset \cite{helber2019eurosat} based on the CLIP model. The label noise is significantly unbalanced among different categories, where the accuracy of 6-th category (``Pasture'') is only 0.01 while the accuracy of 8-th category (``Residential'') is up to 0.95.}
\label{fig:res}
\vskip -0.1in
\end{figure}

To explore robust learning from noisy labels, a series of studies have been conducted, which can be roughly divided into three categories: 1) label correction methods, 2) loss correction methods, and 3) refined training strategies. Label correction methods focus on rectifying noisy labels with the help of complex noise models, i.e., directed graphical models \cite{xiao2015learning} and conditional random fields \cite{vahdat2017toward}.
However, in order to obtain the noise model, the support from extra clean data is indispensable.
The idea of loss correction methods \cite{patrini2017making,han2018masking} is to modify the objective functions for training deep neural networks robustly. It holds the belief that assigning importance weights to examples increases the robustness of the training objective. The methods of refined training strategies \cite{malach2017decoupling,jiang2018mentornet,ma2018dimensionality} promote the robustness of deep neural networks via modifying the training paradigms. For instance, Co-teaching \cite{han2018co} is proposed to maintain two peer networks simultaneously during training, in which one network is backpropagated by the selected confident samples from another network to alleviate the accumulated error.
\begin{figure*}[]
  \centering
  \centerline{\includegraphics[width=12cm]{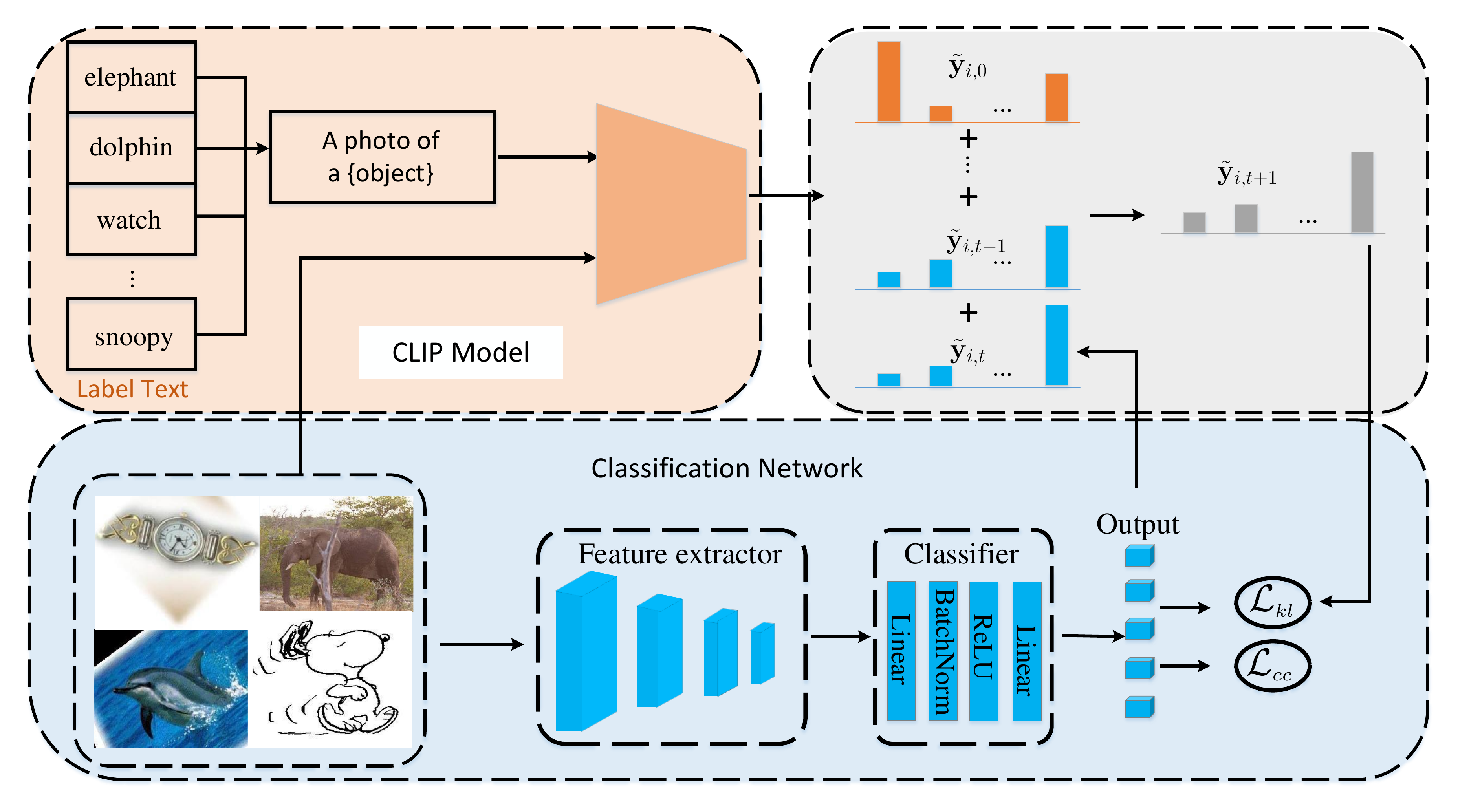}}
\vskip -0.2in
\caption{The pipeline of the proposed method. Here CLIP model is merely required to obtain the label text of each category to generate text embedding so as to annotate the unlabeled images. Modules in blue is trained from scratch with the initial supervision from CLIP model and the training labels are updated in an iterative learning strategy to filter out the label noise. Specifically, top right is the \emph{ensemble labels} module while bottom is the noisy label learning process regularized by \emph{class-conditional contrastive learning} loss. In this paper, we use VIT-B/32 \cite{dosovitskiy2020vit} as the backbone of CLIP model for zero-shot pseudo labeling and use ResNet-50 as the backbone of the classification network for transductive learning.}
\label{fig:res}
\vskip -0.1in
\end{figure*}

To make noisy label learning more trackable, it is required to leverage the intrinsic information of the unlabeled data. For example, the consistency regularization from the semi-supervised learning holds the assumption that the prediction of an instance should not be too different from its perturbation one \cite{sohn2020fixmatch}. Under this inspiration, recent works explore to handle the label noise by integrating the wisdom of semi-supervised learning technology. In order to achieve this goal, DivideMix \cite{li2020dividemix} designs two diverged networks: one network uses the dataset division from another one alternately to separate the clean data (considered as labeled data) and the noisy data (considered as unlabeled data). Then the semi-supervised training are conducted with the improved MixMatch \cite{berthelot2019mixmatch} to perform label co-refinement and label co-guessing on the labeled and unlabeled data, respectively.

Despite the remarkable empirical results achieved by the above mentioned methods, they hold the same assumption: the noisy label is simulated and balanced with known noise rate of each category. As shown in Figure 1, the noisy labels generated from CLIP model violate the above assumption. As a result, an obvious obstacle in the existing methods is the confirmation bias \cite{arazo2020pseudo}: the performance is restricted when learning from severely inaccurate pseudo labels. To escape from the dilemma, we propose in this paper a novel design of Transductive CLIP. Specifically, this paper has the following contributions: 1) This paper proposes to leverage the unlabeled data that supervised from CLIP model in a transductive learning manner to alleviate the burden of data labeling. 2) To tackle confirmation bias problem, a \emph{class-conditional contrastive learning} ($C^3L$) mechanism is proposed to mitigate the reliance on pseudo labels and boost the tolerance to noisy labels. 3) To stabilize the training of deep neural networks, an \emph{ensemble labels} scheme is utilized to update the incorrect pseudo labels in an iterative learning strategy.

\section{Methodology}
\label{sec:format}
We describe the framework in detail as follows. Given a dataset of unlabeled instances $\mathbf{X}=\{\mathbf{x}_i\}_{i=1}^{n}$ with the text name of categories included in the dataset. The initial training labels $\tilde{\mathbf{Y}}_{0} =\{\tilde{\mathbf{y}}_{i,0}\}_{i=1}^{n}$ are generated from CLIP model, where $\mathbf{x}_i \in \mathcal{R}^d$ ($d$ is the dimension of each instance) and $\tilde{\mathbf{y}}_{i,0} \in \mathcal{R}^C$ ($C$ is the number of categories). The goal of this paper is to build a classification network from scratch, which can be trained on the noisy labels robustly and generate higher-quality pseudo labels for the unlabeled data.

\subsection{Class-Conditional Contrastive Learning}
To exploit the knowledge from the initial training labels, it is natural to force the output of the network consistent with the noisy labels. We apply temperature scaling to recalibrate the training labels to reduce the uncertainty of training labels. The training label of $i^{th}$ instance can be reformulated as
\begin{equation}
	(\tilde{\mathbf{y}}_{i,t})_j = \frac{((\tilde{\mathbf{y}}_{i,t})_j)^{\tau}}
{\sum_{j'=1}^{C}((\tilde{\mathbf{y}}_{i,t})_{j'})^{\tau}}
\end{equation}
where $\tau$ is the rescaling factor and is set as $2$ in all experiments (the value of $\tau$ is relatively stable over a large interval). The subscript $j$ denotes the category index. $t$ is the training epoch and the initialization $\tilde{\mathbf{y}}_{i,0}$ of training label $\tilde{\mathbf{y}}_{i,t}$ comes from the output of the CLIP model. The training label will be updated in each training epoch which will be introduced in Sec.\ref{ensemble-labels}. The training objective of model $f$ on the unlabeled data is obtained by minimizing the following Kullback-Leibler divergence loss (${l}_{kl}$),
\begin{equation}
\begin{small}
	\mathcal{L}_{kl}=
	\sum\nolimits_{i=1}^{m} \ \ell_{kl}\left(f_t(\mathbf{x}_i) \ ||\ \tilde{\mathbf{y}}_{i,t}\right),
\end{small}
\end{equation}
where $f_t(\mathbf{x}_i)$  is the output of classification model through a \textbf{Softmax} function.
$m$ is the batch size in the training phase. ${\mathcal{L}}_{kl}$ tends to focus on optimizing the error between network output and training labels, which is effective and efficient in clean-annotated labels. However, the ${\mathcal{L}}_{kl}$ is not robust enough for the noisy labels. Training samples with noisy labels will be stuck into wrong category predictions in the early stage of training, and are difficult to be corrected later due to the memory effect of deep neural network, which is known as confirmation bias issue. When it comes to noisy label, the classification model trained by ${\mathcal{L}}_{kl}$ will be easily confused by false pseudo-labels since it focuses on learning a hyperplane for discriminating each class from the other classes.

To remedy the class discrimination, contrastive learning \cite{chen2020simple,he2020momentum} improves the quality of the learned representations by exploring the intrinsic structure of instances. However, the optimization of standard contrastive learning loss is independent of the training labels, leaving the useful discriminative information on the shelf. To address this problem, we propose to integrate the class discrimination and instance discrimination to cope with the training of noisy label. Consequently, the hyperplane is joint optimized. This strategy aims to uncover the underlying structure of network's output to reduce the overconfidence of the network on its predictions (Softmax output, namely class-conditional scores). Utilizing such discriminative information, the class-conditional contrastive learning ($C^3L$) loss is defined as:
\begin{equation}
\mathcal{L}_{cc}=
-\sum_{i=1}^{m}\log \frac{\exp \left(\operatorname{sim}\left(f_t(\mathbf{x}_i), f_t(\tilde{\mathbf{x}}_i)\right) / T\right)}{\sum_{k=1}^{m} \exp \left(\operatorname{sim}\left(f_t(\mathbf{x}_i), f_t(\tilde{\mathbf{x}}_k)\right) / T\right)}
\end{equation}
where the $\operatorname{sim}\left(,\right)$ is selected as the cosine function and $T$ is the temperature parameter following the setting in contrastive learning. $f_t(\tilde{\mathbf{x}}_i)$ is the \textbf{Softmax} output of the model with the augmented input $\tilde{\mathbf{x}_i}$,
which implicitly encodes the normalized distance between the instance feature and the learnable class prototypes. The $C^3L$ loss $\mathcal{L}_{cc}$ maximizes the similarity of category prediction between differently augmented views of the same data point. According to the properties of the Softmax function adopted in equation (3), the similarity of category prediction between different data point is minimized, which could push category prediction away from noisy labels. The optimization of the predicted probability will lead to the optimization of the hyperplane.  The clean label will win this prediction competition, since their prediction are easier to fit and achieve higher score, compared to that of false ones. Therefore, the model can not cause serious over-fitting in the training of pseudo-labels, which greatly alleviates the problem of confirmation bias. The overall training loss is 
\begin{equation}
\mathcal{L}_{total}=\mathcal{L}_{kl}+\lambda\mathcal{L}_{cc}
\end{equation}
$\lambda$ is set to 1 by default in this paper.

\subsection{Noisy Labels Rectification with Noise Filtering}
\label{ensemble-labels}
The training labels with false predictions tend to fluctuate. For example, the predictions of one sample may be predicted as one object in one epoch and another object in another epoch, which usually contains higher label noise. To obtain reliable prediction on unlabeled examples and improve the quality of training labels, we adopt the \emph{ensemble labels} scheme to update the training label in an iterative learning strategy. In particular, the training labels are updated by:
\begin{equation}
\tilde{\mathbf{y}}_{i,(t+1)} \leftarrow \frac{\tilde{\mathbf{y}}_{i,0} +\sum_{j=1}^{t}f_j(\mathbf{x}_i)}{t+1}, \forall \mathbf{x}_i \in\left\{\mathbf{x}_i\right\}_{i=1}^{n}
\end{equation}
thus the training labels with higher label noise can be suppressed in this way. During the label updating progress, note that the predictions in different training epochs contribute equally which avoids negative updating.

\section{EXPERIMENTS}
\begin{table*}
\centering
\caption{Classification accuracies (\%). VIT-B/32 (CLIP Backbone) $\to$ Resnet-50 (Target Backbone).}
\begin{tabular}{lrrrrr  >{\columncolor{mygray}} r}
\toprule
Methods   & Caltech101  & DTD   & EuroSAT & NWPU-RESISC  & Flower102 & Average\\
\midrule
CLIP Model& 81.33       & 43.12  & 30.77   & 54.83        & 64.90 & 54.99\\
\midrule
FixMatch  & 82.09       & 43.17  & 33.00   & 60.50        & 66.61 & 57.07\\
Re-weighting
          & 83.56       & 44.46  & 33.53   & 60.42        & 67.36 & 57.87\\
Dash      & 82.91       & 44.07  & 34.46   & 60.83        & 68.56 & 58.17\\
DivideMix & 80.51       & 45.51  & 37.38   & 70.77        & 70.72 & 60.98\\ \midrule
Ours       & 86.29       & 50.83  & 45.76   & 72.00       & 74.25 & 65.83\\
Gain      & {\color{red}+4.96}       & {\color{red}+7.71}  & {\color{red}+14.99}  & {\color{red}+17.17}      & {\color{red}+9.35} & {\color{red}+10.84} \\
\bottomrule
\end{tabular}
\label{tab:digits}
\vskip -0.14in
\end{table*}
\subsection{Datasets}\noindent\textbf{Caltech101} \cite{fei2006one} consists of images from 101 object categories and a background class from Google search. Each category contains 31 to 800 images with medium resolution, which is around 300$\times$300 pixels. \textbf{Describable Textures Dataset (DTD)}  \cite{cimpoi14describing} is a texture database collected from Google and Flickr. It consists of 5640 images, which are organized according to a list of 47 categories inspired from human perception.  \textbf{EuroSAT} \cite{helber2019eurosat} consists of 27,000 labeled images with 10 different land use and land cover classes.  \textbf{NWPU-RESISC} \cite{cheng2017remote} is a benchmark for Remote Sensing Image Scene Classification (RESISC), created by Northwestern Polytechnical University (NWPU). This dataset contains 31,500 images, covering 45 scene classes with 700 images in each class. \textbf{Flower102} \cite{nilsback2008automated} is an image classification dataset consisting of 102 flower categories.

\subsection{Baseline Methods}
We evaluate the effectiveness of our proposed framework by re-implementing other state-of-the-art methods on our proposed experimental settings:
\begin{itemize}[leftmargin=*]
\item \textbf{FixMatch} \cite{sohn2020fixmatch} is a state-of-the-art semi-supervised learning method which retains the pseudo-label of weakly-augmented unlabeled images when the prediction of model is higher than the confidence-threshold (0.95 by default).
\item \textbf{Re-weighting} is a popular loss correction method in noisy label learning to reduce the label noise and the instance weight is set as the maximum classification probability.
\item\textbf{Dash} \cite{xu2021dash} is a state-of-the-art method which dynamically selects instance whose loss value is less than a dynamic threshold at each optimization step to train the models.
\item\textbf{DivideMix} \cite{li2020dividemix} is a state-of-the-art noisy label learning method. At each mini-batch, one network performs an improved Mixmatch method with the clean set and noisy set that are dynamically divided by another diverged network.
\end{itemize}
\subsection{Implementation Details}
\noindent \textbf{Network architecture.}
In this paper, we employ the VIT-B/32 \cite{dosovitskiy2020vit} as the backbone module of CLIP model and select the ResNet-50 model pre-trained on ImageNet as the backbone module for transductive optimization. Note that VIT-B/32 is substantially larger than ResNet-50. The classifier module is constructed by a two-layer MLP in conjunction with a batchnorm layer as shown in Figure 2. \\
\noindent \textbf{Network hyper-parameters.}
Following the common setting in contrastive learning, the temperature $T$ is adopted as 0.07. The learning rate of the classifier is set to 0.01 and the batch size is set to 128 in all experiments. The learning rate of the classifier is 10 times as the backbone following the common fine-tuning principle. Stochastic gradient descent (SGD) with a momentum of 0.9 is adopted to optimize the model.
\subsection{Performance Analysis}
\noindent  Firstly, the classification accuracies of the proposed method and the four baseline methods on the five datasets are shown in Table 1. We observe that the proposed method achieves much better performance than the baseline methods with statistical significance. The classification accuracy of our method on NWPU-RESISC is 72.00\% and the performance improvement is 17.17\% compared to the CLIP model. Note that, this task is very difficult since most of the baseline methods can only achieve slight improvement and may perform very poorly on many of the datasets. Secondly, in Figure 3 (a), we conduct contrastive learning strategy on the intermediate feature \cite{chen2020simple} to show the superiority of the proposed class-conditional contrastive learning. Compared with standard contrastive learning loss, $C^3L$ loss can achieve higher performance when label noise is high (for example, object with class id 38), which is inseparable from the contribution of model category information during contrastive regularization. Thirdly, we compare our proposed \emph{ensemble labels} strategy with other training label updating strategies, and the result is shown in Figure 3 (b). Even if the training label is not updated (namely \emph{CLIP labels}), the performance of the model can reach around 70 \% on the NWPU-RESISC dataset. This shows that $C^3L$ can handle noisy labels well. For \emph{pseudo labels}, we have
$\tilde{\mathbf{y}}_{i,(t+1)} \leftarrow f_t(\mathbf{x}_i), \forall \mathbf{x}_i \in\left\{\mathbf{x}_i\right\}_{i=1}^{n}$. It updates training labels according to the prediction in the previous epoch, which induces negative updating. Compared with other strategies, the proposed \emph{ensemble labels} can achieve higher performance and a more stable training process.
\begin{figure}
  \centering
\vskip -0.05in
  \centerline{\includegraphics[width=7.5cm]{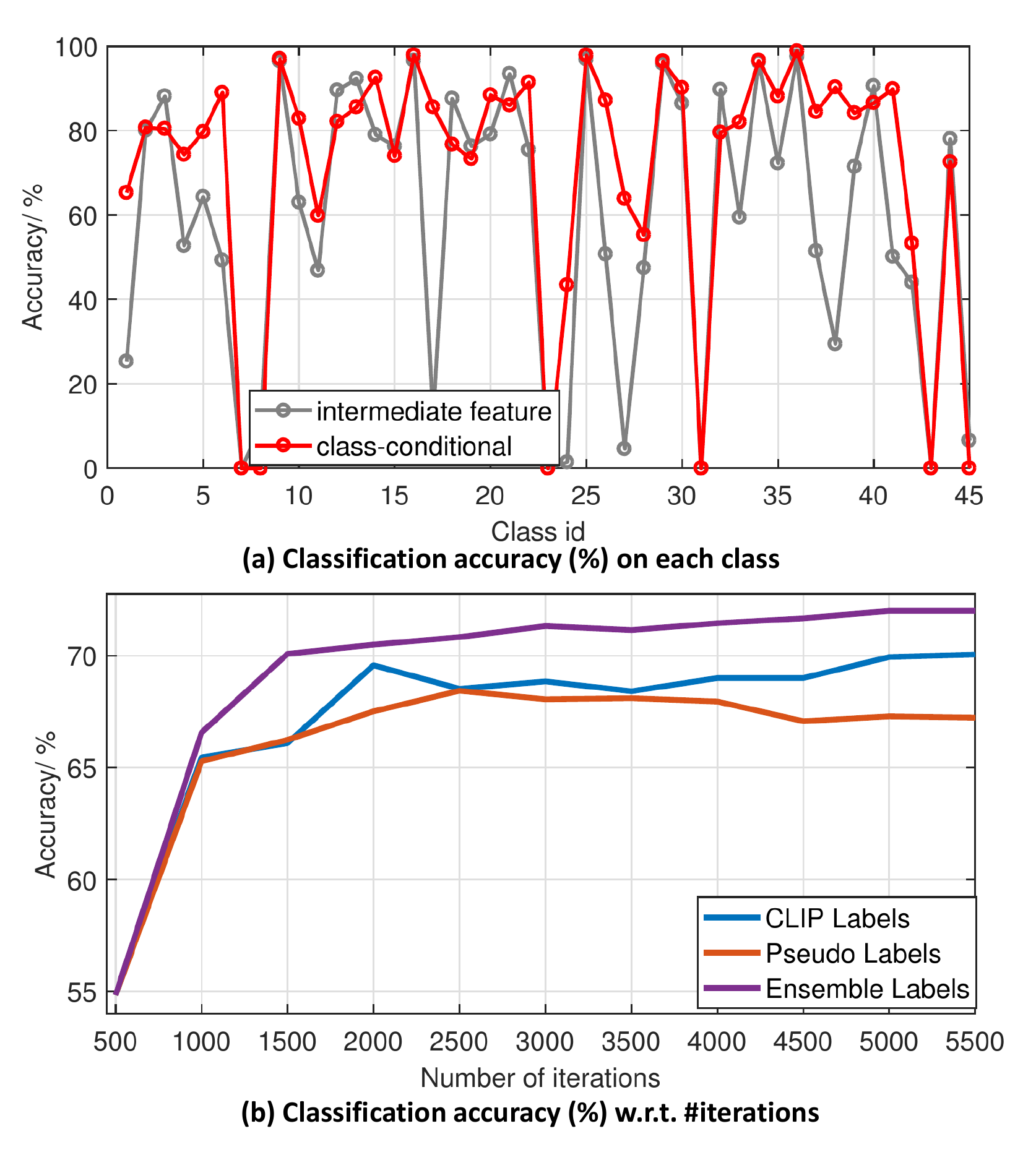}}
\vskip -0.2in
\caption{Effectiveness verification.}
\label{fig:res}
\vskip -0.2in
\end{figure}
\section{CONCLUSION}
To alleviate the burden of data labeling, in this paper we explore an interesting, realistic but challenging task where only the vision-language pre-trained model is provided to the unlabeled data as the supervision. To handle label noise, we propose a simple yet effective framework called Transductive CLIP. As a component of the proposed method, the \emph{class-conditional contrastive learning} loss integrates the class discrimination and instance discrimination to mitigate the reliance on pseudo labels and boost the tolerance to noisy labels. Furthermore, we adopt the \emph{ensemble labels} scheme to update the training label in an iterative learning strategy to rectify noisy labels. Extensive results on five image classification tasks verify the efficacy of our proposed method.
\vfill\pagebreak
\label{sec:refs}
\small
\bibliographystyle{IEEEbib}
\bibliography{c3l}
\end{document}